\documentclass[journal]{IEEEtran}
\usepackage{amsmath,amsfonts}
\usepackage{algorithmic}
\usepackage{algorithm}
\usepackage{array}
\usepackage[caption=false,font=normalsize,labelfont=sf,textfont=sf]{subfig}
\usepackage{textcomp}
\usepackage{stfloats}
\usepackage{url}
\usepackage{verbatim}
\usepackage{graphicx}
\usepackage{cite}
\usepackage{multirow}
\usepackage{booktabs}
\usepackage[colorlinks=true, linkcolor=blue, citecolor=blue, urlcolor=black, filecolor=blue]{hyperref}
\begin{document}

\title{FedPylot: Navigating Federated Learning for Real-Time Object Detection in Internet of Vehicles}

\author{Cyprien Quéméneur,~\IEEEmembership{Graduate Student Member,~IEEE,} and Soumaya Cherkaoui,~\IEEEmembership{Senior Member,~IEEE}
\thanks{This work was supported by the Natural Sciences and Engineering Research Council of Canada (NSERC). (\emph{Corresponding author: Cyprien Quéméneur.})}
\thanks{The authors are with the Department of Computer and Software Engineering, Polytechnique Montréal, Montreal, QC H3T 1J4 Canada (e-mail: cyprien.quemeneur@polymtl.ca; soumaya.cherkaoui@polymtl.ca).}
}

\maketitle

\begin{abstract}
The Internet of Vehicles~(IoV) emerges as a pivotal component for autonomous driving and intelligent transportation systems~(ITS), by enabling low-latency big data processing in a dense interconnected network that comprises vehicles, infrastructures, pedestrians and the cloud. Autonomous vehicles are heavily reliant on machine learning~(ML) and can strongly benefit from the wealth of sensory data generated at the edge, which calls for measures to reconcile model training with preserving the privacy of sensitive user data. Federated learning~(FL) stands out as a promising solution to train sophisticated ML models in vehicular networks while protecting the privacy of road users and mitigating communication overhead. This paper examines the federated optimization of the cutting-edge YOLOv7 model to tackle real-time object detection amid data heterogeneity, encompassing unbalancedness, concept drift, and label distribution skews. To this end, we introduce FedPylot, a lightweight MPI-based prototype to simulate federated object detection experiments on high-performance computing~(HPC) systems, where we safeguard server-client communications using hybrid encryption. Our study factors in accuracy, communication cost, and inference speed, thereby presenting a balanced approach to the challenges faced by autonomous vehicles. We demonstrate promising results for the applicability of FL in IoV and hope that FedPylot will provide a basis for future research into federated real-time object detection. The source code is available at \url{https://github.com/cyprienquemeneur/fedpylot}.
\end{abstract}


\section{Introduction} \label{introduction}

\begin{figure}[!t]
    \centering
    \includegraphics[width=\linewidth]{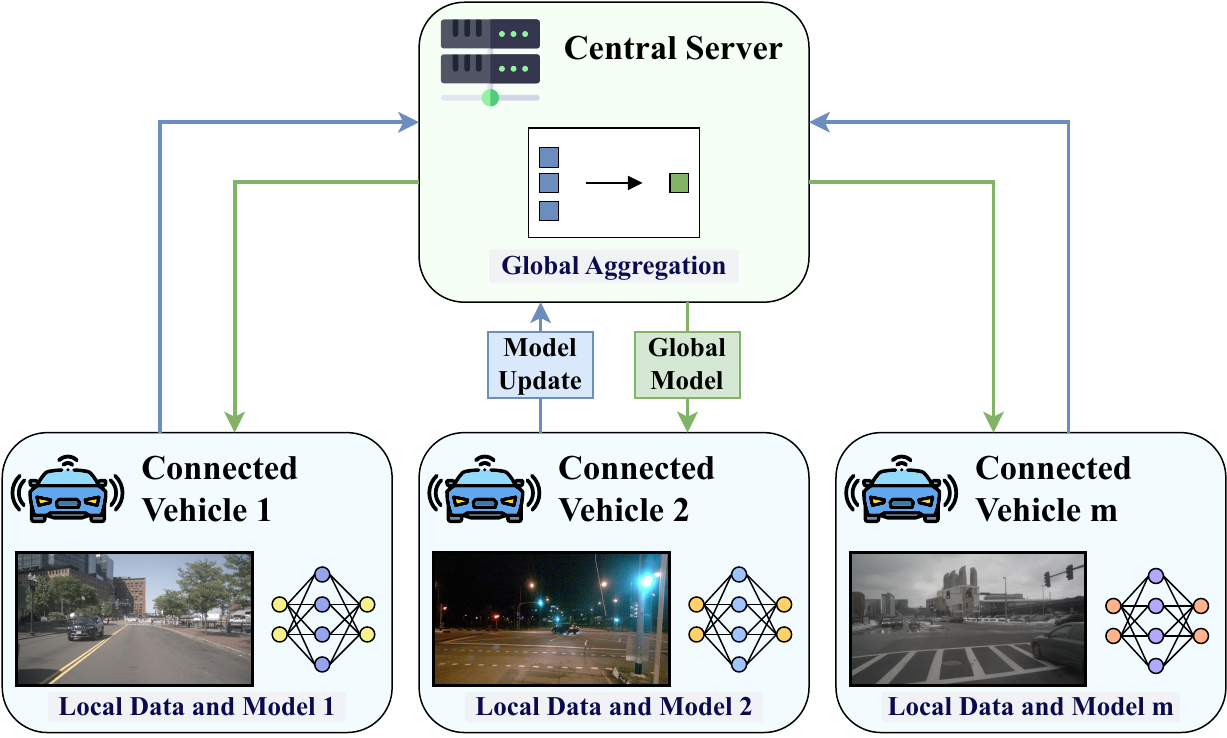}
    \caption{\textbf{Vehicular clients collaboratively learn a joint model with FL}. The vehicles collect driving data using various sensors to train their own local model, while a central server is responsible for regularly gathering and aggregating local weight-update vectors to compute a global model, which is subsequently disseminated to the clients for further training. The raw data are kept private, but are typically non-identically distributed due to the decentralized nature of the clients (samples taken from nuImages).}
    \label{fig_1}
\end{figure}

\IEEEPARstart{I}{ntelligent} transportation systems (ITS) are expected to reshape mobility by enhancing safety, streamlining traffic flow, reducing vehicle emissions and fuel consumption, and providing infotainment services. This transformation is powered by advances in machine learning (ML) and vehicle-to-everything (V2X) communication technologies, fostering seamless cooperation between a network of vehicles, pedestrians, and infrastructures, generating a vast amount of data and integrated into a cohesive Internet of Vehicles (IoV)~\cite{xuInternetVehiclesBig2018}. To enable data sharing, IoV relies on state-of-the-art wireless network technologies that can offer long-range, low latency, reliable, and secured transfers~\cite{storckSurvey5GTechnology2020}. In turn, connected automated vehicles can leverage the information sharing facilitated by IoV to enhance their situational awareness, yet their abilities to solve advanced navigation tasks nonetheless hinge on machine learning~(ML) models~\cite{grigorescuSurveyDeepLearning2020a}. Furthermore, although vehicles can utilize an array of sensors such as cameras, LiDAR, radar, and GPS systems to collect diverse multimodal data, crucial for supporting ML and making timely decisions, the offloading of model training to the cloud raises privacy, availability, and, in the long run, scalability concerns. To palliate these issues, federated learning~(FL) has been proposed as a solution to facilitate collaborative edge model training and protect user privacy while alleviating the communication bottlenecks arising in IoV~\cite{luBlockchainEmpoweredAsynchronous2020, manias_making_2021}. In FL, raw data sharing is prohibited and training takes place locally on edge clients, which then rely on a central aggregation agent to regularly gather and combine model updates. This process is depicted for IoV in Fig.~\ref{fig_1}, where the main clients are vehicles, and the central aggregation server is strategically positioned at a network edge location to reduce latencies. Additionally, edge servers dispersed across a given geo-region may themselves fallback on a cloud platform to facilitate orchestration at scale, or provide a second layer of aggregation~\cite{zhou_two-layer_2021}.

Real-time vision is an indispensable requirement for automated vehicles, as attaining full driving autonomy involves sophisticated vision-based systems capable of achieving human-level perception in complex and dynamic environments~\cite{janai_computer_2020}. Numerous studies have explored the potential to improve the visual perception capabilities of vehicles with FL, and covered traffic sign recognition~\cite{xieEfficientFederatedLearning2022, padariaTrafficSignClassification2023, lianTrafficSignRecognition2024}, pothole and other road damage detection~\cite{yuanFedRDPrivacypreservingAdaptive2021, alshammari3PodFederatedLearningbased2022, sahaFederatedLearningBased}, semantic segmentation~\cite{fantauzzo_feddrive_2022, fani_feddrive_2023}, and more commonly object detection~\cite{rjoub_improving_2021, chen_federated_2021, bommel_active_2021, rjoub_active_2022, dai_online_2023, wangFederatedDeepLearning2023, chi_federated_2023, rao_sparse_2023, su_cross-domain_2023,kim_navigating_2023, zheng_autofed_2023, mishra_swarm_2023, chi_federated_2023-1, khalilFederatedLearningHeterogeneous2024, khalilDrivingEfficiencyAdaptive2024, jallepalli_federated_2021}. Other tasks relying in part but not solely on the visual perception of vehicles have also been explored with FL, such as trajectory prediction~\cite{hanFederatedLearningbasedTrajectory2022} and collision avoidance~\cite{yuPersonalizedDrivingAssistance2022}. However, these investigations typically employ either models not suitable for real-time vision or relevant but outdated ML models. This, in turn, diminishes the applicability of the findings for evaluating the feasibility of FL with modern real-time vision ML models. In contrast, this work shifts its focus to YOLOv7~\cite{wang_yolov7_2023}, one of the latest entries in the YOLO (You Only Look Once) series of real-time object detection models. Through numerous successive improvements, YOLO models have gained significant recognition in the field of computer vision and have been widely used in various applications, including autonomous vehicles, surveillance systems, and robotics~\cite{tervenComprehensiveReviewYOLO2023a}. Furthermore, this research explores the application of YOLOv7 within a FL framework for autonomous driving situations marked by data sources stemming from vehicles associated with different geographical locations and timeframes, resulting in vehicles encountering heterogeneous data.

The contributions of this paper can be summarized as follows:
\begin{itemize}
    \item We develop FedPylot, a Message Parsing Interface (MPI)-based FL prototype dedicated to federated object detection experiments on high-performance computing (HPC) systems, and implement a hybrid cryptosystem
    to secure the communications between FL participants.
    \item We propose, to the best of our knowledge, the first federated framework for YOLOv7 that allows for high-scale experimentation. We considered predictive performances, inference speed and communication overhead in our evaluation, emphasized local optimization and included server-side momentum and custom learning rates.
    \item We demonstrate FedPylot on two relevant autonomous driving datasets and simulate data heterogeneity arising from spatiotemporal shifts, and account for unbalancedness, concept drift, and label distribution skews. We capture heterogeneity at different granularity by including two class maps of a long-tail distribution, while considering navigation sequences in our splitting strategy.
\end{itemize}
\noindent We make FedPylot open-source in the hope that it will be helpful to the object detection community for their FL-related projects. FedPylot retains the ability to scale to a large number of computation nodes, while being easier to approach than advanced FL frameworks (e.g., FedML~\cite{he_fedml_2020}, PySyft~\cite{zillerPySyftLibraryEasy2021}, Flower~\cite{beutelFlowerFriendlyFederated2022a}), in which integrating complex self-defined models, like state-of-the-art object detectors, can prove troublesome.

The remainder of this paper is organized as follows. Section~\ref{background} outlines the fundamental concepts and tools used in this work. Subsequently, Section~\ref{related_work} provides a review of the current literature surrounding the intersection of object detection, federated learning, and autonomous driving. In Section~\ref{system_design}, we detail the theoretical design of our FL prototype. Section~\ref{experiments} is dedicated to the implementation of FedPylot and our experimental settings. Section~\ref{results} outlines the results of our experiments. Finally, Section~\ref{conclusion} concludes the paper.
\section{Background} \label{background}

\subsection{Federated learning}

\subsubsection{Formulation}

FL was proposed in 2016 by McMahan et al.~\cite{mcmahan_ce_2017}. In traditional FL, \(m\) clients \(\{\mathcal{C}_{1}, \mathcal{C}_{2}, ..., \mathcal{C}_{m}\}\), with the help of an orchestrating server \(\mathcal{S}\), participate during several communication rounds in the collaborative training of a shared machine learning model \(w\) without revealing their respective local dataset \(\{\mathcal{D}_{1}, \mathcal{D}_{2}, ..., \mathcal{D}_{m}\}\). We assume a horizontal partitioning where the local datasets of the clients differ in their data samples but share the same feature space. The size of each local dataset is designated by \(n_{i}=|\mathcal{D}_{i}|\), with \(n=\sum_{i=1}^{m} n_{i}\). The optimization goal is to minimize
\begin{equation}
\label{eq1}
    \operatorname*{min}_w \mathcal{L}(w) = \sum_{i=1}^{m} \frac{n_{i}}{n} L_{i}(w)
\end{equation}

\noindent where \(L_{i}(w) = \mathbb{E}_{(x, y) \sim \mathcal{D}_{i}} [\ell_{i}(x, y ; w)]\) is the possibly non-convex local objective function of client \(\mathcal{C}_i\). At the beginning of the training process, the central server initializes the global model from random or pre-trained weights and shares it with the clients. In the original and baseline algorithm FedAvg, at the beginning of the communication round \(t\), a subset of clients with indices in \(K\) is randomly selected to participate in the round. These clients then receive the shared model \(w^{t}\) from the server and perform several epochs \(E\) of local training on their respective local dataset using stochastic gradient descent~(SGD) with local mini-batch size \(B\). The central server then collects the updated local models \(w_{i}^{t}\) from each client \(\mathcal{C}_i\) and aggregates them, thus yielding the next instance of the global model
\begin{equation}
\label{eq2}
    w^{t+1} = \sum_{i \in K} \frac{n_{i}}{n} w_{i}^{t}.
\end{equation}

\noindent Reddi et al.~\cite{reddi_adaptive_2021} formalized FedOpt as a direct generalization of FedAvg. In FedOpt, it is no longer assumed that the local optimizers of the clients are necessarily SGD, and the update rule of the global model is rephrased as an optimization problem. Assuming that \(\mathcal{C}_i\) transmits the weight-update vector \(\Delta_{i}^{t} = w^{t} - w_{i}^{t}\) to the server, the latter can aggregate the local updates to form a \emph{pseudo-gradient} \(\Delta^{t}\), which is then inputted to a server optimizer. In particular, (\ref{eq2}) is equivalent to
\begin{align}
\label{eq3}
    w^{t+1} = w^{t} - \Delta^{t} &= w^{t} - \sum_{i \in K} \frac{n_{i}}{n} \Delta_{i}^{t}
\end{align}

\noindent which corresponds to using SGD as the server optimizer with a learning rate of 1. To assess the convergence of the training procedure, it is necessary to evaluate the global model. Evaluation can be performed server-wise on a separate dataset \(\mathcal{D}_{\mathcal{S}}\) left on the server, or separately on some clients, before aggregating the resulting local statistics. The original FL process can be declined in a variety of ways, for example, personalized federated learning (pFL) allows for some degree of customization of the shared model that is no longer unique to all clients~\cite{tan_towards_2023}, while some implementations leverage blockchain to achieve complete decentralization by eliminating the need for a central server entirely~\cite{nguyen_federated_2021}; one may also assume a vertical partitioning, where the local datasets differ in the feature space instead of the sample space~\cite{yang_federated_2019}.

\subsubsection{Data heterogeneity}
In FL, the data are usually non-identically distributed (non-IID) among the clients, which may lead to local model divergence during training and degraded accuracy for the global model. Kairouz et al. ~\cite{kairouz_advances_2021} presented various forms of data heterogeneity, three of which are relevant to our study:
\begin{itemize}
    \item Concept drift: The same label may have vastly different features for different clients, for example, due to varying weather conditions, locations, or time scales.
    \item Label distribution skew: Differences in the distribution of data labels can arise across clients, for example, because the likelihood of encountering certain labels is tied to some specific clients' environment.
    \item Unbalancedness: The number of data samples stored can vary greatly from client to client.
\end{itemize}
FL practitioners have devised several strategies to simulate data heterogeneity in their experiments. In image classification, label skewness is widely studied and the Dirichlet distribution is commonly used to create artificial non-IID splits of a dataset~\cite{hsu_measuring_2019}. Conversely, in tasks such as object detection, data heterogeneity is multifaceted and better simulated by identifying natural semantic separations within the dataset, such as weather or location in the case of ITS. Many popular techniques have been introduced to address the issue of data heterogeneity, including, but not limited to, FedProx~\cite{li_federated_2020} which introduces a proximal term to each local objective function to improve robustness against variable local updates, SCAFFOLD~\cite{karimireddy_scaffold_2020} which uses control variates to perform variance reduction and counterbalance the client-drift, and MOON~\cite{li_model-contrastive_2021} which applies contrastive learning at the model level to correct local training based on similarities between model representations. However, many of these algorithms were originally designed for image classification and may not be as effective when applied as is to more challenging vision tasks~\cite{miaoFedSegClassHeterogeneousFederated2023b}. Other methods are more readily compatible with object detection, and we discuss them in the following. Strategies based on replacing server-side SGD by a more advanced ML optimizer, such as Adam~\cite{kingma_adam_2017}, are orthogonal to the ones mentioned above and can improve convergence in non-IID settings~\cite{reddi_adaptive_2021}. Similarly, Hsu et al.~\cite{hsu_measuring_2019} introduced momentum in server-side optimization to create FedAvgM and empirically showed improved performances under heterogeneous distributions. In FedAvgM, the global update integrates a velocity term \(v\), which accumulates exponentially decaying past pseudo-gradients at a rate controlled by a constant momentum factor \(\beta \in [0, 1)\). FedAvgM can be generalized by adding a supplementary constant learning rate \(\eta\), the update rule thus becoming as follows
\begin{subequations}\label{eq4}
\begin{align}
v^{t+1} &= \beta v^{t} + \Delta^{t} \label{eq4A}\\
w^{t+1} &= w^{t} - \eta v^{t+1} \label{eq4B}
\end{align}
\end{subequations}

\noindent and where taking \(\beta=0\) and \(\eta=1\) yields back the original FedAvg algorithm. Momentum can be applied at the client-level as well to improve the stability of local updates, while retaining convergence in non-IID settings~\cite{liu_accelerating_2020, xu_fedcm_2021}. Starting federated learning after a pre-training phase instead of random initialization, using proxy data available on the server, has also been shown to improve the stability of global aggregation and close the gap with centralized learning, even when data are heterogeneous~\cite{chenImportanceApplicabilityPreTraining2022, nguyenWhereBeginImpact2022}.

\subsubsection{Other challenges in FL} \label{FL_challenges}
FL elicits a multitude of other challenges, such as designing incentives to encourage the clients with the largest data pools to participate in the federated process, ensuring fairness between clients, accounting for heterogeneity in the computing capabilities of the clients, designing FL frameworks, and limiting overheads. Furthermore, despite its added privacy benefits, FL is not exempt from risks. Malicious actors participating can gain insight into the original training data or degrade training integrity by conducting attacks such as model inversion and model update poisoning. Counter techniques like secure multi-party computation~\cite{yao_how_1986}, homomorphic encryption~\cite{gentry_fully_2009}, and differential privacy~\cite{dwork_calibrating_2006} can help alleviate this problem, but may incur increased computation and communication overheads or performance degradation.

\subsection{Object detection}

Object detection is a computer vision task in which a model seeks to detect all instances of object classes of interest in an image and report their location and spatial extent by delimiting them with bounding boxes~\cite{zaidi_survey_2022}. In real-time object detection, the inference speed of the model is also deemed critical to the realization of the task and is commonly measured either in milliseconds or in frames-per-second (FPS). Deep learning made a substantial impact to object detection in the past decade, and modern object detectors are usually based on convolutional neural networks. Two-stage object detectors, such as R-CNN~\cite{girshick_rich_2014} and SPP-Net~\cite{he_spatial_2015}, first generate region proposals before sending them to a classification model, while one-stage object detectors, such as YOLO~\cite{redmon_you_2016}, SSD~\cite{liu_ssd_2016} and RetinaNet~\cite{linFocalLossDense2017}, locate and classify objects in a single swipe, usually making them faster but less accurate. One-stage and two-stage object detectors alike commonly rely on the non-maximum suppression (NMS) post-processing technique to filter overlapping predictions and retain only the most relevant bounding box for a given object. More recently, transformer-based object detectors have gained interest, such as detection transformers (DETRs)~\cite{carion_end--end_2020}. DETRs eliminate the need for many hand-designed components such as NMS to perform end-to-end detection, and efforts are being directed to enable them in the real-time setting~\cite{lv_detrs_2023}.

The most popular metrics used to evaluate the predictive performance of object detectors are reviewed by Padilla et al.~\cite{padilla_survey_2020}, for which we propose a brief recapitulation in the following. The location accuracy of the predicted bounding box \(B_{p}\) at threshold \(t\) is given by the ratio between the area of overlap and the area of union of \(B_{p}\) and the ground truth \(B_{gt}\), and is called the intersection over union
\begin{equation}
\label{eq5}
    \text{IoU} = \frac{|B_{p} \cap B_{gt}|}{|B_{p} \cup B_{gt}|}
\end{equation}

\noindent with the prediction being deemed correct if \(\text{IoU} \geq t\). Having established the criterion to determine the correctness of the detection, it is possible to derive the precision \(\text{P}=\frac{\text{TP}}{\text{TP}+\text{FP}}\) and recall \(\text{R}=\frac{\text{TP}}{\text{TP}+\text{FN}}\) of the model, where TP, FP and FN refer to True Positive, False Positive and False Negative, respectively. The average precision \(\text{AP}^{k}_{t}\) of the model for class \(k\) at threshold \(t\) is then interpolated from the area under curve (AUC) of the \(\text{precision} \times \text{recall}\) curve. The overall accuracy of the model evaluated on a dataset of \(N\) classes is called the mean average precision, and is defined as
\begin{equation}
\label{eq6}
    \text{mAP}_{t} = \frac{1}{N} \sum_{k=1}^{N}{\text{AP}^{k}_{t}}.
\end{equation}

\noindent To benchmark a model using a single threshold \(t\) may not be satisfactory. Therefore, it is common to average the mean average precision across ten IoU thresholds, from 50\% to 95\% with a step of 5\%, resulting in a metric which is simply referred to as mAP.

\subsection{YOLOv7}

YOLOv7 is one of the latest entries in the YOLO family of one-stage real-time object detectors. At release, in July 2022, it was the fastest and most accurate object detector in the 5 to 160 FPS range, had fewer parameters than comparatively performing models, and was a leap from the still broadly popular YOLOv5~\cite{Jocher_YOLOv5_by_Ultralytics_2020}. YOLOv7 was trained from scratch on MS COCO~\cite{lin_microsoft_2014} and features optimized structures and optimization methods dubbed \emph{trainable bag-of-freebies}. These include planned re-parameterized convolution based on gradient flow path analysis, and a novel label assignment strategy, as well as some pre-existing concepts, like embedding batch normalization statistics directly in convolutional layers for inference, YOLOR implicit knowledge \cite{wangYouOnlyLearn2023} merged into convolutional layers for additions and multiplications, and using an exponential moving average (EMA) model as the final inference model. Bag-of-freebies strengthens the training cost and predictive performance of a model but without increasing latencies during inference, a concept previously discussed in YOLOv4~\cite{bochkovskiy_yolov4_2020}. YOLOv7 also introduced architectural improvements, including Extended-ELAN, a modified variant of the Efficient Layer Aggregation Network (ELAN) \cite{wangDesigningNetworkDesign2022a} that allows stacking computational blocks indefinitely while retaining learning ability, and a new method for compound scaling of concatenation-based models that preserves the model's properties and optimal structure.

More generally, YOLOv7 is an anchor-based model that uses a feature pyramid network (FPN)~\cite{lin_feature_2017}. The authors introduce YOLOv7-tiny and YOLOv7, which are P5 models respectively designed for edge and normal GPUs, and YOLOv7-W6, which is a P6 model designed for cloud GPUs. The compound scaling method is applied to YOLOv7 to derive YOLOv7-X, and to YOLOv7-W6 to derive YOLOv7-E6, YOLOv7-D6, YOLOv7-E6E, with the latter replacing ELAN by Extended-ELAN as its main compute unit. The recommended training input resolution is \(640\times640\) for P5 models and \(1280\times1280\) for P6 models. Images passed to YOLOv7 are resized to the given input while maintaining aspect ratio, while padding is applied if necessary. During training, the loss is computed by summing three sub-functions balanced by predefined gain hyperparameters. These sub-functions measure the performance of the model across different modalities of the object detection problem. In particular, the classification loss and objectness loss use the common Binary Cross-Entropy (BCE) loss, whereas the bounding box regression loss sub-function is instead based on the Complete IoU (CIoU) loss~\cite{zheng_distance-iou_2020}. All YOLOv7 variants use the SiLU activation function (except for YOLOv7-tiny where LeakyReLU is used instead), the SimOTA strategy introduced in YOLOX for label assignment~\cite{geYOLOXExceedingYOLO2021a}, mosaic, mixup and left-right flip augmentations, gradient accumulation, automatic mixed precision training, half precision at inference, and NMS for post-processing.

In addition to 2D object detection, YOLOv7 was extended to support pose estimation and instance segmentation, and an anchor-free variant, YOLOv7-AF, was made available with base performances on par or surpassing later releases, including YOLOv6 3.0~\cite{li_yolov6_2023} and YOLOv8~\cite{Jocher_Ultralytics_YOLO_2023}. YOLOv7 was used as the basis for YOLOv9 \cite{wangYOLOv9LearningWhat2024}, which introduced Programmable Gradient Information (PGI) and the Generalized Efficient Layer Aggregation Network (GELAN).
\section{Related work} \label{related_work}

In the following, we propose a brief review of previous work that applied FL to tackle 2D object detection in autonomous vehicles. Rjoub et al.~\cite{rjoub_improving_2021} investigated federated real-time object detection in adverse weather driving scenarios with the original YOLO model. Chen et al.~\cite{chen_federated_2021} studied FL in vehicular object detection with SSD under communication and computation constraints. Bommel~\cite{bommel_active_2021} and later Rjoub et al.~\cite{rjoub_active_2022} used active learning, with respectively YOLOv5s and YOLOR, to address the predicament of sparse data labeling in FL for real-time object detection in autonomous vehicles. Dai et al.~\cite{dai_online_2023} proposed FLAME, a framework intended to facilitate the exploration of online federated object detection on data continuously streamed by autonomous vehicles, which they demoed with YOLOv2~\cite{redmonYOLO9000BetterFaster2017a}. Wang et al.~\cite{wangFederatedDeepLearning2023} designed CarlaFLCAV, an open FL simulation platform that supports a wide range of automotive perception tasks, including object detection with YOLOv5, and tackled network resource and road sensor placement optimization. Chi et al.~\cite{chi_federated_2023} leveraged a soft teacher semi-supervised object detection framework to perform FL training with Faster-R-CNN~\cite{renFasterRCNNRealTime2017b} on unlabeled data collected while driving, given a small amount of well-curated preexisting data. Rao et al.~\cite{rao_sparse_2023} proposed FedWeg, a sparse FL training process that they evaluated on YOLOv3~\cite{redmonYOLOv3IncrementalImprovement2018a}, to accommodate computational and communication constraints in IoV. Su et al.~\cite{su_cross-domain_2023} proposed FedOD, a cross-domain pFL framework for object detection based on multi-teacher distillation, and validated their proposal with RetinaNet on autonomous driving datasets. Finally, Kim et al.~\cite{kim_navigating_2023}  proposed a two-stage training strategy named FedSTO and tackled semi-supervised federated object detection with YOLOv5 in heterogeneous situations where the local datasets of the vehicular clients are fully unlabeled and labeled data are only available to the central server. 

In contrast, other researchers instead shifted their attention to multimodal data. In particular, Zheng et al.~\cite{zheng_autofed_2023} proposed AutoFed, a FL framework dedicated to bird's-eye view vehicle detection where data heterogeneity results from the inclusion of multiple sensing modalities, and trained a custom two-stage object detector that accommodates LiDAR and radar data on several NVIDIA Jetson TX2 devices. Mishra et al.~\cite{mishra_swarm_2023} argued in favor of a fully decentralized blockchain-based autonomous driving FL system with smart contracts, called \emph{swarm learning} in the paper, and validated their proposal on 3D point cloud object detection with Complex-YOLOv4~\cite{simonComplexYOLOEulerRegionProposalRealTime2019}. Moreover, Chi et al.~\cite{chi_federated_2023-1} showed that federated 3D object detection can be improved by incorporating infrastructure into a clustered federated procedure, and included a complex multistep perception system in their experiments, where feature maps extracted from point clouds were processed by a vision transformer.

Our proposal diverges from previous work by focusing on the federated optimization of a recent real-time object detector and proposing a comprehensive evaluation of its performances, while allowing high-scale simulations in data centers. Our splitting strategy, which we detail in Section~\ref{splitting_strategy}, features particularities not found in these papers. We also propose one of few works to provide an open implementation.

Two papers experimented with YOLOv8 and respectively introduced FedProx+LA, a FL method to address label distribution skews in vehicular networks which showed improved performance and convergence speed for object detection~\cite{khalilFederatedLearningHeterogeneous2024}, and an adaptive clustered FL technique to address storage and bandwidth limitations, validated for car detection under varying weather and lighting conditions~\cite{khalilDrivingEfficiencyAdaptive2024}. However, these works only considered the nano, i.e., the smallest, variant of YOLOv8, which is designed for edge devices and has limited learning capabilities. This further emphasizes the need to provide support for larger-scale FL experimentation. Perhaps most similar to our proposal is the prototype introduced by Jallepalli et al.~\cite{jallepalli_federated_2021} which, to the best of our knowledge, is the only contribution that has featured the federated optimization of an object detector in an HPC environment for the purpose of autonomous driving. We significantly improve upon this prior work by replacing YOLOv3 by YOLOv7, introducing higher data diversity, more clients, realistic non-IID settings, more configurations for local and server-wise optimization, measurements of communication costs and inference speed for several model variants, and by replacing low-level socket-based communications by MPI, and the symmetric Fernet encryption scheme by a more secure hybrid one.
\section{System design} \label{system_design}

\begin{figure}[!t]
    \centering
    \includegraphics[width=\linewidth]{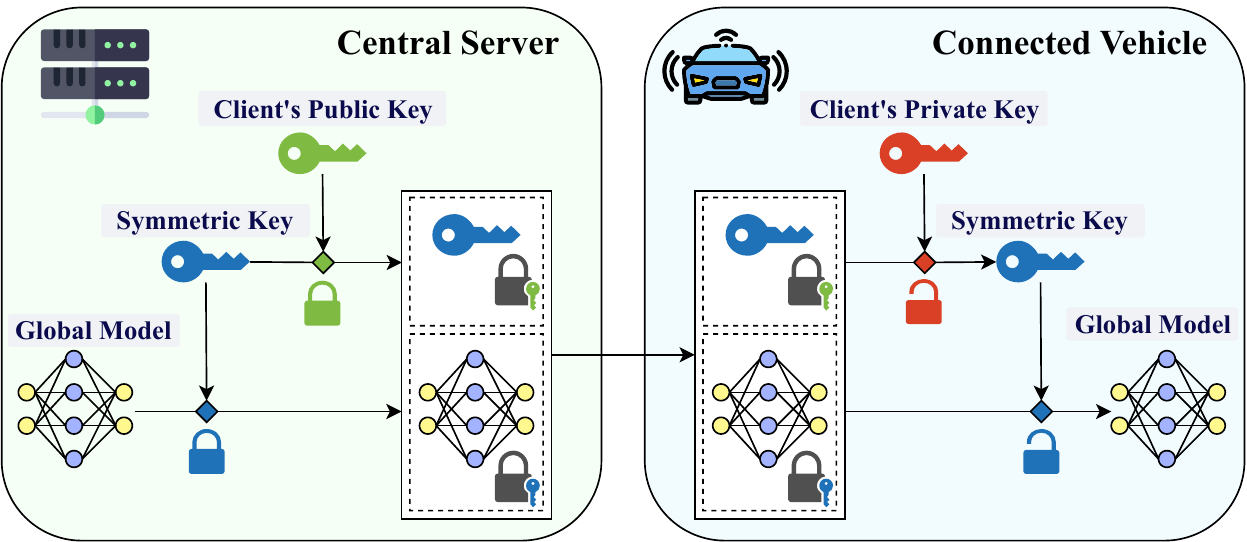}
    \caption{\textbf{Hybrid cryptosystem for server-client communications}. Transmissions between the server and the vehicular clients are encrypted using a highly efficient symmetric algorithm. The symmetric key is generated by the server and protected using a public-key cryptosystem when passed to the clients.}
    \label{fig_2}
\end{figure}

\begin{algorithm}[!t]
\caption{Federated YOLOv7}
\label{alg:alg1}
  \begin{algorithmic}[0]
    \vspace{0.1cm}
    \STATE \textbf{Input:} Communication rounds \(R\), local epochs \(E\), local mini-batch size \(B\), server learning rate \(\eta\), server momentum factor \(\beta \in [0, 1)\), YOLOv7 set of hyperparameters \(\mathcal{H}\).
    \vspace{0.1cm}
    \STATE \textbf{Output:} Best re-parameterized joint model \(w_{\text{r}}^{*}\).
    \\\hrulefill
    \vspace{0.1cm}
    \STATE \textbf{Server \(\mathcal{S}\) executes:}
    \begin{ALC@g}
    \STATE Gather public key \(pk_{i}\) from each client \(\mathcal{C}_i\)
    \STATE Initialize model \(w^{0}\) from pretrained weights
    \STATE Initialize server-side momentum \(v^{0}=0\)
    \FOR{\(t = 0, ..., R-1\)}
        \STATE Initialize symmetric key \(sk^{t}\)
        \STATE \(\{w^{t}\}\leftarrow\) Encrypt \(w^{t}\) with \(sk^{t}\)
        \STATE \(\{sk^{t}\}_{i:1:m}\leftarrow\) Encrypt \(sk^{t}\) for each \(pk_{i}\)
        \FOR{each client \(\mathcal{C}_i\) \textbf{in parallel}} 
            \STATE\(\{\Delta_{i}^{t}\}\), \(n_{i} = \) \textbf{ClientUpdate}(\(\{w^{t}\}, \{sk^{t}\}_{i}, t\))
        \ENDFOR
        \STATE \(n = \sum_{i=1}^{m} n_{i}\)
        \STATE \(\Delta_{i:1:m}^{t} \leftarrow\) Decrypt each \(\{\Delta_{i}^{t}\}\) with \(sk^{t}\)
        \STATE \(\Delta^{t} = \sum_{i=1}^{m} \frac{n_{i}}{n}\Delta_{i}^{t}\)
        \STATE \(v^{t+1} = \beta v^{t} + \Delta^{t}\)
        \STATE \(w^{t+1} = w^{t} - \eta v^{t+1}\)
        \STATE \(w_{\text{r}}^{t+1} = \texttt{reparameterize}(w^{t+1})\)
        \STATE \(\text{Evaluation metrics} = \texttt{test}(w_{\text{r}}^{t+1}, \mathcal{D}_{\mathcal{S}}, \mathcal{H})\)
    \ENDFOR
    \STATE \textbf{return} \(w_{\text{r}}^{*}\) based on mAP
    \end{ALC@g}
    \vspace{0.1cm}
    \STATE \textbf{ClientUpdate(\(\{w^{t}\}, \{sk^{t}\}_{i}, t\)):} \hspace{0.2cm} // \emph{run on client \(\mathcal{C}_i\)} 
    \begin{ALC@g}
        \STATE \(sk^{t} \leftarrow\) Decrypt \(\{sk^{t}\}_{i}\) with private key
        \STATE \(w^{t} \leftarrow\) Decrypt \(\{w^{t}\}\) with \(sk^{t}\)
        \STATE \(w_{i}^{t} = \texttt{train}(w^{t}, \mathcal{D}_{i}, E, B, \mathcal{H}, t)\)
        \STATE \(\Delta_{i}^{t} = w^{t} - w_{i}^{t}\)
        \STATE \(\{\Delta_{i}^{t}\} \leftarrow\) Encrypt \(\Delta_{i}^{t}\) with \(sk^{t}\)
        \STATE \textbf{return} \(\{\Delta_{i}^{t}\}\), \(|\mathcal{D}_{i}|\) to server
    \end{ALC@g}
  \end{algorithmic}
\label{alg1}
\end{algorithm}

Several connected and autonomous vehicles collect video data in real time using onboard cameras, while participating in the federated training of a shared object detection model using the YOLOv7 architecture. A server acts as the FL system orchestrator and is responsible for the collection and aggregation of local model updates, server-side optimization, and dissemination of model parameters. The server possesses its own set of well-curated representative data, which is used to assess the quality of the global model at the end of each communication round. The server optimizer used in our experiments is FedAvgM~(\ref{eq4}), although FedPylot also supports FedAdagrad, FedAdam and FedYogi~\cite{reddi_adaptive_2021}. To more accurately reflect the state of the model in deployment, evaluation is performed on the re-parameterized model, which has fewer parameters yet retains the same predictive performances as the base model. A round encompasses a fixed number of local training epochs common to all clients with a shared batch size irrespective of the size of the training sets, hence the number of local optimization steps may vary between clients. For simplicity, we assumed full client participation for each round, availability of the training data labels at the edge, and synchronicity of aggregation.

Communications between participants are secured through a hybrid cryptosystem. It consists in combining a highly efficient symmetric encryption algorithm, which is used to encrypt the plaintext but requires the same symmetric key to be available to both the sender and the receiver, with a more practical but several orders of magnitude more expensive public-key encryption algorithm, used only to protect the symmetric key from adversaries when it is being transferred. We design our system so that each vehicular client generates a public-private key pair at the beginning of the federated process and immediately transmits its public key to the server. Meanwhile, the server generates a new symmetric key at the beginning of each round and uses it to encrypt the global model, before passing the two to the clients, either alongside each other as in Fig.~\ref{fig_2}, or separately if decoupling key and model exchanges is deemed more practical. The clients can then use this same key to decrypt the global model and encrypt their local updates. Hybrid encryption is straightforward and well adapted to environments with low-latency constraints. Indeed, it incurs only negligible communication overhead and, in our setting, the encryption overhead is largely dominated by the cost of model training. However, it is only suitable when revealing the model updates to the server is acceptable. If this condition is not met, the addition of advanced privacy techniques, such as those mentioned in Section~\ref{FL_challenges}, becomes mandatory. The federated procedure is summarized in Algorithm~\ref{alg:alg1}, while the remaining details are covered in the next Section.
\section{Experiments} \label{experiments}

\subsection{Prototype implementation details}

\subsubsection{Federated simulation}
Our experiments were carried out with PyTorch~\cite{paszke_pytorch_2019} on Cedar.\footnote{Cedar is a general-purpose computer cluster of the Digital Research Alliance of Canada. Documentation is at \url{https://docs.alliancecan.ca/wiki/Cedar}.} Each FL participant was identified with exactly one compute node equipped with one Tesla~V100-SXM2-32GB GPU, eight CPU cores, and up to 6000~MiB of memory per core. Before the beginning of a FL experiment, the local dataset of each client was transferred from the network storage to the local disk of the corresponding compute node, isolating it from the other participants, and accelerating input and output (I/O) transactions. The same was done with the validation set and the server node. The interconnect was Intel Omni-Path~\cite{birrittellaIntelOmnipathArchitecture2015}, and we handled the communications between the server and the clients with MPI, using the Open MPI implementation of the standard~\cite{gabriel_open_2004}, and the mpi4py Python package~\cite{dalcin_mpi4py_2021}. MPI was specifically designed for HPC environments and is available as a communication backend in several FL frameworks, including FedML. During a simulation, one MPI process is created per compute node, i.e., per FL participant, and the central server uses collective communications, including \emph{gathering}, \emph{broadcasting}, and \emph{scattering}, to orchestrate transfers. In practice, we distinguish the first round, where the entire model checkpoint initialized from pre-trained weights must be transmitted to the clients, from the following rounds, where only the transmission of the learnable parameters is required. Furthermore, weights were represented using half-precision~(FP16) during transfer to reduce communication overhead.

\subsubsection{Encryption}
The hybrid cryptosystem used to secure the communications between the FL participants was implemented with Python's cryptography package. The key encapsulation scheme is based on RSA~\cite{rivest_method_1978} with the OAEP~\cite{bellareOptimalAsymmetricEncryption1995} padding scheme, while data encapsulation was handled with a 256-bit symmetric key using the Advanced Encryption Standard (AES) algorithm~\cite{daemen_aes_1999} with Galois/Counter Mode (GCM)~\cite{mcgrew_galoiscounter_2004}, and with scrypt as the password-based key derivation function~\cite{percival_stronger_2009}. AES-GCM requires the use of a cryptographic nonce for each new encryption to prevent replay attacks. In our scheme, the 12-byte nonce used during each data encryption and the salt used when creating a new symmetric key were all generated with Python's secrets package, to ensure the use of cryptographically strong random numbers. Encrypting with AES-GCM also yields a 16-byte authentication tag, which can be safely transmitted alongside the nonce and ciphertext to ensure the integrity and authenticity of the message.

\subsection{Datasets}

\begin{figure}[!t]
    \centering
    \includegraphics[width=\linewidth]{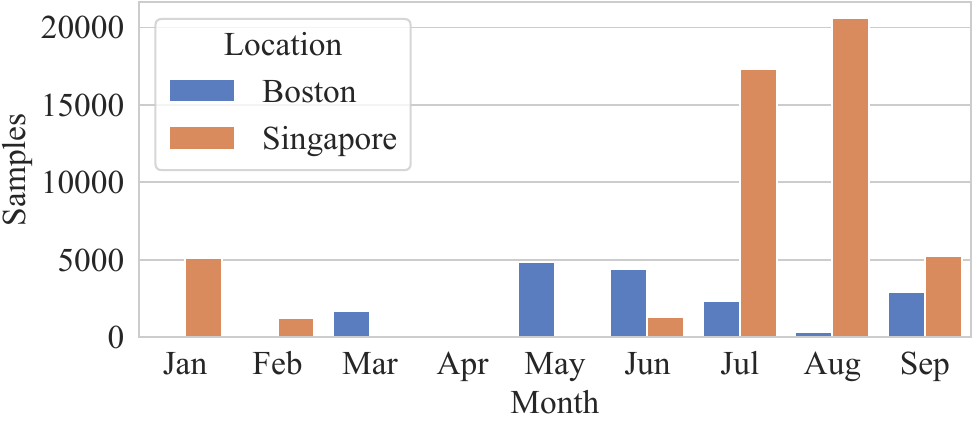}
    \caption{\textbf{Distribution of samples in nuImages training data.} The dataset is composed of nearly 500 driving logs acquired through six different cameras, and features a wide range of driving scenarios and weather conditions}
    \label{fig_3}
\end{figure}

We conducted experiments on two relevant autonomous driving datasets: the 2D object detection subset of the KITTI Vision Benchmark Suite~\cite{geiger_are_2012} and nuImages, an extension of nuScenes dedicated to 2D object detection~\cite{caesar_nuscenes_2020}. KITTI is a pioneering and still widely popular benchmark, which features synchronized stereo RGB images, GPS coordinates, and LiDAR point clouds, and supports 2D and 3D object detection. However, it suffers from low data diversity in terms of weather and lightning conditions, location, and object orientation. The 2D object detection subset of KITTI does not feature a pre-existing train/validation split and contains 7481 labeled images, with the most common dimension being \(1242\times375\) (small variations are present). KITTI contains a \emph{DontCare} class corresponding to regions where objects were not labeled which was excluded from our experiments, thus leaving eight classes in the dataset. nuImages contains almost 500 driving logs of varying length and is organized as a relational database, featuring 67279 labeled training images and a separate predefined validation set of 16445 labeled images, all samples being of size \(1600\times900\). The data were acquired through six cameras oriented to provide a 360-degree view around the vehicle with some small overlaps. The scenes were taken in Boston and Singapore from January to September, as shown in Fig.~\ref{fig_3}, and feature rain, snow, and nighttime, making nuImages significantly more diverse than KITTI. nuImages also includes 23 different classes spread in a long-tail distribution. We did not consider the predefined non-annotated test sets of KITTI and nuImages in our experiments. In both cases, the bounding boxes are represented with their top left and bottom right absolute coordinates, thus the annotations of the bounding boxes first had to be converted to the YOLO format, where a bounding box is represented with normalized center coordinates, box width, and height. 

\subsection{Data-splitting strategy} \label{splitting_strategy}

\begin{figure*}[!t]
    \centering
    \includegraphics[width=\linewidth]{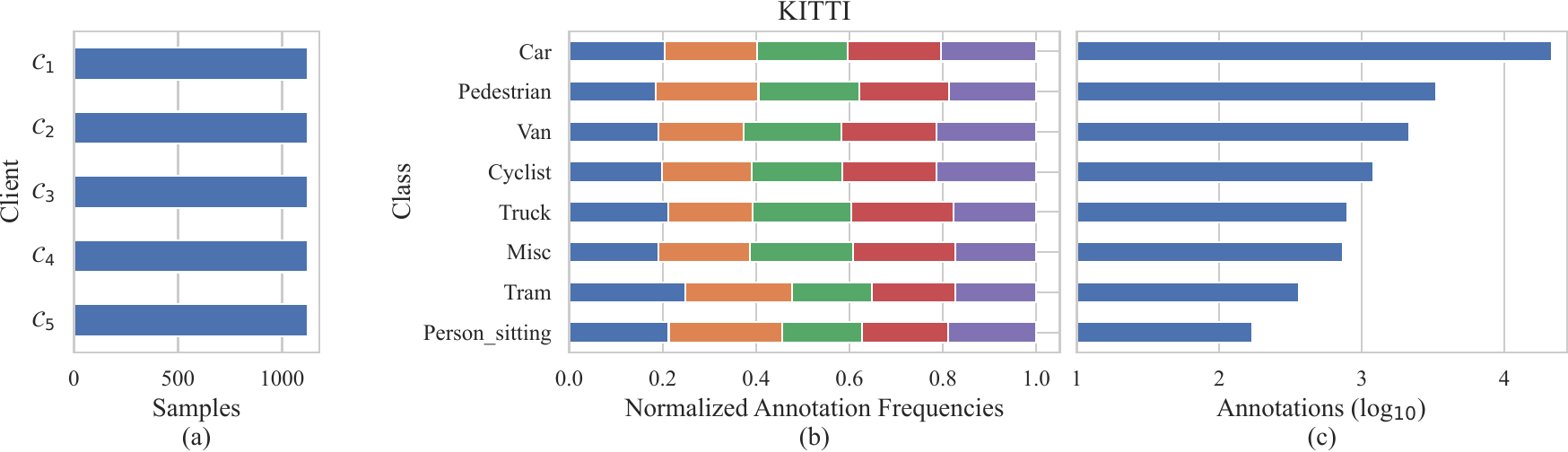}
    \caption{\textbf{KITTI split.} 25\% of KITTI training data are stored on the server, while the 75\% left are distributed IID among five clients. We report (a) the number of samples held by each client, (b) the normalized number of annotations held per sample per client, and (c) the label distribution in the original training set.}
    \label{fig_4}
\end{figure*}

\begin{figure*}[!t]
    \centering
    \includegraphics[width=\linewidth]{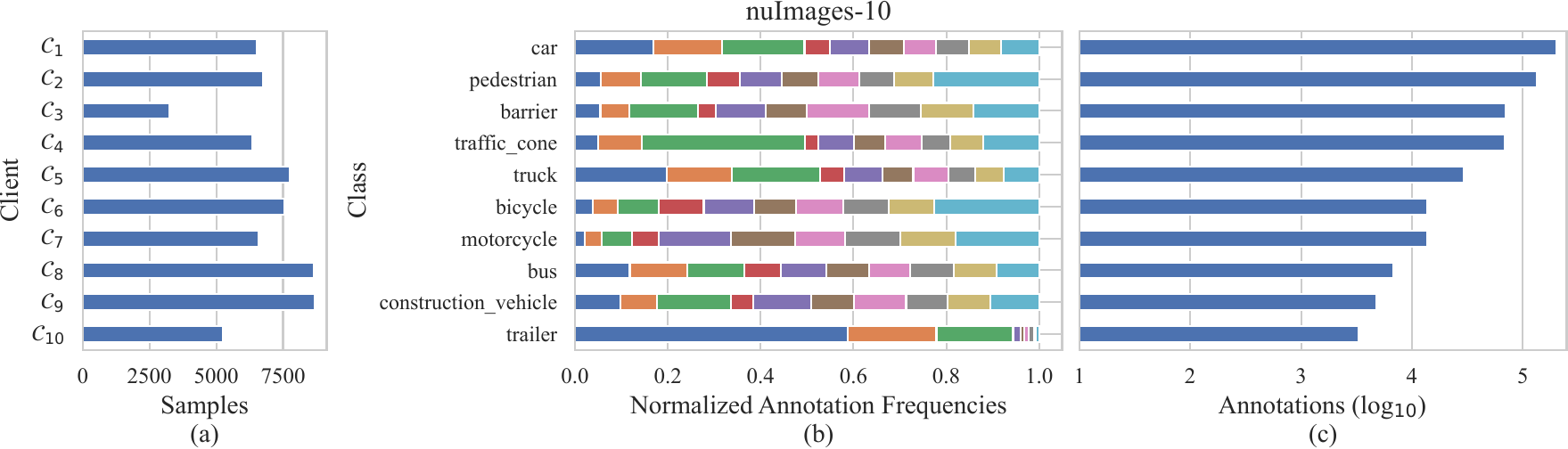}
    \caption{\textbf{nuImages-10 split.} The original classes are mapped to ten labels, and the training data are split non-IID among ten clients. We report (a) the number of samples held by each client, (b) the normalized number of annotations held per sample per client, and (c) the label distribution in the original training set.}
    \label{fig_5}
\end{figure*}

\begin{figure*}[!t]
    \centering
    \includegraphics[width=\linewidth]{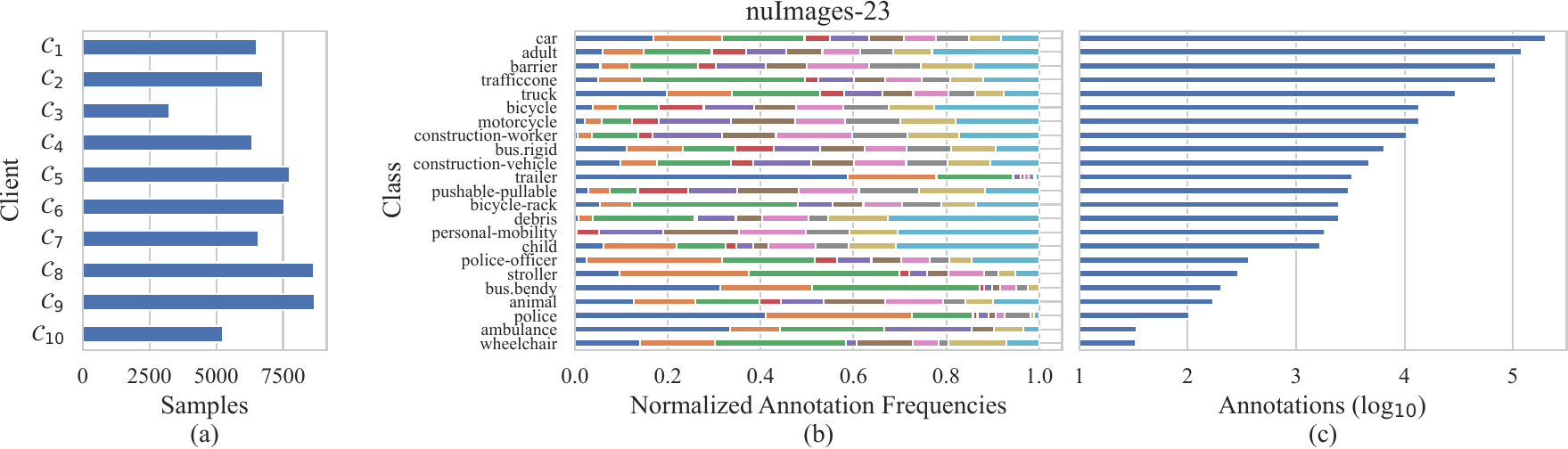}
    \caption{\textbf{nuImages-23 split.} The full long-tail distribution is retained, and the training data are split non-IID among ten clients. We report (a) the number of samples held by each client, (b) the normalized number of annotations held per sample per client, and (c) the label distribution in the original training set.}
    \label{fig_6}
\end{figure*}

As KITTI is a low-variance dataset, which does not allow for any easy natural semantic separation of the data, it served as our IID setting. We randomly sampled 25\% of the training data to store on the central server, while the remaining samples were distributed homogeneously among five clients, as shown in Fig.~\ref{fig_4}. Regarding nuImages, we created non-IID splits by relying on the spatiotemporal metadata available for the training set to generate unbalancedness, concept drifting, and label distribution skewness. The set held by the server is simply constituted of nuImages's predefined validation data, while the training splits are organized as follows. Clients \(\mathcal{C}_{1}\), \(\mathcal{C}_{2}\) and \(\mathcal{C}_{3}\) received the data collected in Boston, respectively, for the months of March and May, June and July, and September, while \(\mathcal{C}_{4}\) received the data collected in Singapore in January and February. Singapore data from June to August, which are overrepresented in the training set, were distributed among clients \(\mathcal{C}_{5}\) to \(\mathcal{C}_{9}\); however, the distribution was not made based on individual samples, but by randomly sampling entire data logs, thus ensuring a minimal threshold of data heterogeneity based on the specificities of each navigation sequence. Finally, client \(\mathcal{C}_{10}\) received the data of September from Singapore. Furthermore, we implemented two class maps for nuImages to observe the impact of the dataset's long-tailed distribution on federated optimization. We refer to the dataset resulting from the map replicated from the nuScenes competition, where only ten classes are retained, as nuImages-10, and to the base dataset with all 23 classes as nuImages-23. We note that the aforementioned splitting strategy naturally leads to unbalancedness and label distribution skews, as shown in Fig.~\ref{fig_5} and Fig.~\ref{fig_6}, with the latter being further exacerbated when the full long-tail distribution is included. In particular, among the clients differing only by their navigation sequences, the differences in label distributions only become stark for the rarer classes, thus illustrating how our splits capture heterogeneity at varying granular levels.

\subsection{Design of experiments}

For the first set of experiments, we implemented FedAvg as a baseline. The local and server-side optimizers were both SGD with a fixed learning rate of respectively 0.01 and 1.

The second FL method, FedOpt, focuses on client-side optimization. The server-side optimizer is still SGD, but we adapted the original YOLOv7 training procedure to the federated setting, with default fine-tuning hyperparameters. Specifically, the model parameters are divided into three groups consisting of the biases, the batch normalization parameters, and the remaining parameters to which weight decay is applied with 0.0005 chosen as the unscaled regularization constant. Each group adopts its own one-cycle learning rate scheduling policy, which includes a linear warm-up period followed by cosine decay annealing. The learning rate is initialized at 0.1 for the bias group and at 0 for the two remaining groups, and all the learning rates evolve towards 0.01 during warm-up, and 0.001 during decay. To maintain synchronicity, we fixed the warm-up duration to be a fixed number of epochs common to all clients (respectively 30 and 15 for KITTI and nuImages). Furthermore, Nesterov momentum is also applied locally in FedOpt, with a starting momentum factor of 0.8 which is linearly increased to 0.937 during warm-up and maintained constant thereafter. We chose not to aggregate the local momentums, thus reducing communication overhead but forcing the clients to maintain a persistent state between rounds of training. A possible improvement to FedPylot would be to implement an additional stateless variant of client-level momentum, to support it irregardless of the participation rate.

Lastly, we performed ablation with the FedAvgM (\ref{eq4}) server-side optimizer alongside the local optimization described in FedOpt, a combination which we refer to as FedOptM. We performed a grid search for the server learning rate \(\eta \in \{0.5, 1.0, 1.5\}\) as well as the momentum factor \(\beta \in \{0.1, 0.3, 0.5, 0.7, 0.9\}\). We restricted \(\eta\) to a small region around its default value, as the local learning rate schedules were left unchanged.

In all instances above, we used the original anchor-based YOLOv7 architecture with weights pre-trained on MS COCO, letterbox resize of 640, a batch size of 32, mosaic and horizontal flipping augmentations with respective probabilities 1.0 and 0.5, a confidence threshold of 0.001, an IoU threshold for NMS of 0.65 on testing data, and gains of respectively 0.05, 0.3 and 0.7 for the box regression, classification, and objectness losses. When applicable, we allowed the learning rates and the momentum factor to evolve within the communication rounds to accommodate the communication constraints in IoV. Local EMA models were not aggregated and the scheduling policies were maintained locally. FedAvg and FedOpt were compared against each other during 150 epochs of training and with rounds of different length, \((R, E) \in \{(30, 5), (15, 10), (10, 15)\}\), whereas for FedOptM we specifically focused on the 30 rounds setting to allow for more server optimization steps. FL was compared against the default centralized procedure, where the model was trained out of the box on all data for 150 epochs. For simplicity, the warm-up length was kept the same as in the federated experiments.
\section{Results} \label{results}

\subsection{Effects of client-side optimization}

\begin{table}[!t]
\caption{Impact of client-side optimization on YOLOv7 testing accuracy compared against centralized learning\label{tab:table1}}
\centering
\begin{tabular}{cccccc}
\toprule
\multirow{2}{*}{Dataset}     & \multirow{2}{*}{Centralized} & \multicolumn{4}{c}{Federated}     \\ \cmidrule(lr){3-6}
& & Method   & R30E5  & R15E10 & R10E15 \\
\midrule
\multirow{2}{*}{KITTI} & \multirow{2}{*}{73.7\%} & FedAvg & 57.5\% & 59.5\% & 59.2\% \\
& & FedOpt & 66.3\% & 67.4\% & \textbf{68.3\%} \\
\midrule
\multirow{2}{*}{nuImages-10} & \multirow{2}{*}{52.9\%} & FedAvg & 45.2\% & 45.4\% & 45.2\% \\
& & FedOpt & 47.5\% & \textbf{47.8}\% & 47.3\% \\
\midrule
\multirow{2}{*}{nuImages-23} & \multirow{2}{*}{33.2\%} & FedAvg & 26.6\% & 26.5\% & 26.2\% \\ 
&  & FedOpt & 28.2\% & \textbf{28.3}\% & 28.1\% \\
\bottomrule
\end{tabular}
\end{table}
\begin{figure*}[!t]
    \centering
    \includegraphics[width=\linewidth]{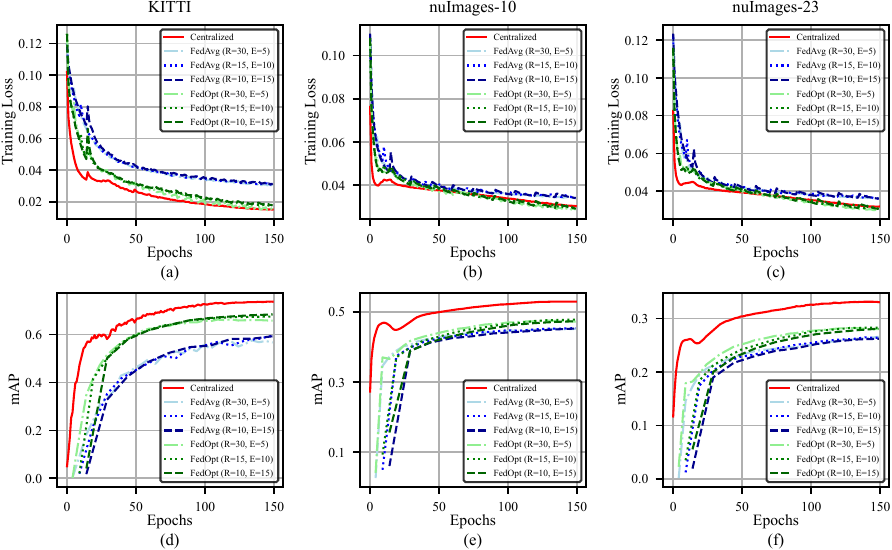}
    \caption{\textbf{Client-side optimization results}. Three round lengths were considered for the federated baseline FedAvg and the locally optimized algorithm FedOpt. We report the evolution for the centralized and federated settings of YOLOv7's (a) training loss on KITTI, (b) training loss on nuImages-10, (c) training loss on nuImages-23, (d) testing accuracy on KITTI, (e) testing accuracy on nuImages-10, and (f) testing accuracy on nuImages-23.}
    \label{fig_7}
\end{figure*}

\begin{figure*}[!t]
    \centering
    \includegraphics[width=\linewidth]{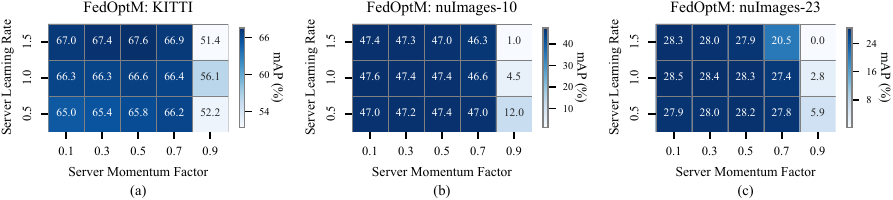}
    \caption{\textbf{Grid search on FedOptM server-side hyperparameters}. Training lasted 30 rounds of 5 epochs on (a) KITTI, (b) nuImages-10, and (c) nuImages-23.}
    \label{fig_8}
\end{figure*}

The evolution of the training loss and mAP for centralized learning, as well as FedAvg and FedOpt, is shown in Fig.~\ref{fig_7}, and we report the highest testing accuracy achieved for each setting in Table~\ref{tab:table1}. In the federated setting, the aggregated training loss is derived from the weighted average of the local losses computed by each client following (\ref{eq1}), and the mAP is measured by evaluating the global model at the end of each communication round on a separate set of unseen examples stored on the central server, as defined in Algorithm~\ref{alg:alg1}. The advanced local optimization scheme resulted in a consistent performance increase across all levels of heterogeneity, but more so on the IID setting, where improved local updates are not countered by the client-drift. Longer training rounds were beneficial on IID data, but also did not lead to degradations in the heterogeneous settings, which we attribute to our choice of splitting strategy and use of pre-trained weights. Performance drops, relatively to the centralized setting, were more notable for nuImages-23 than nuImages-10, confirming that the inclusion of the long-tail distribution adversely impacted federated optimization.

\subsection{Ablation on server learning rate and momentum}

\begin{figure*}[!t]
    \centering
    \includegraphics[width=\linewidth]{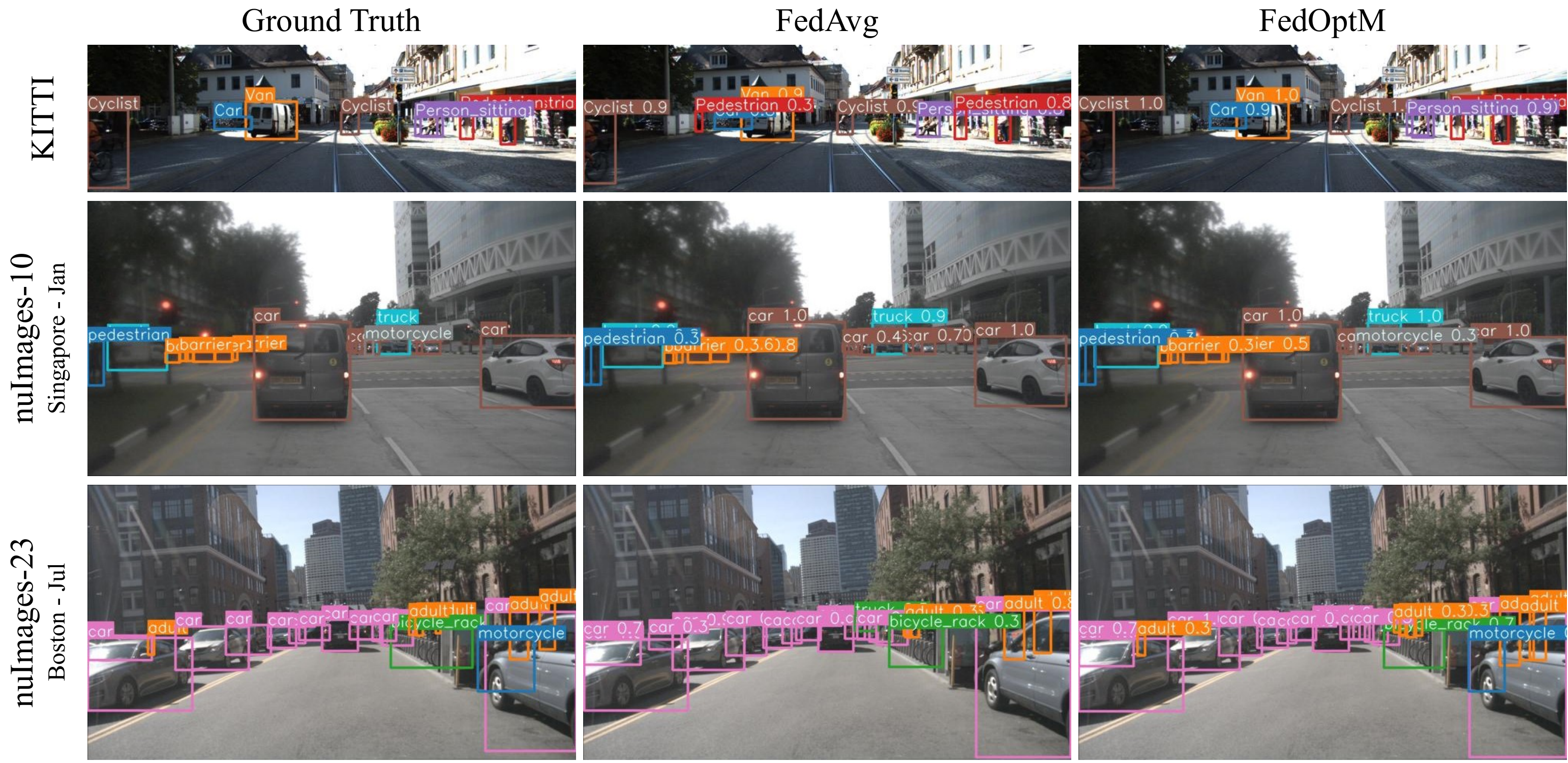}
    \caption{\textbf{Qualitative results of YOLOv7 on testing data.} Images are resized to the input resolution of 640px before inference, while retaining aspect ratio. The model was trained with the FedAvg baseline and the FedOptM optimized procedure on KITTI (IID) and two separate class maps of nuImages (non-IID).}
    \label{fig_9}
\end{figure*}

The full grid search details for FedOptM on the effect of the server learning rate and momentum on testing accuracy are reported in Fig.~\ref{fig_8}, and we comment on the improvements relative to FedOpt for comparable communication rounds number and length. Training on KITTI being very stable, it benefited from higher server-side learning rates, and the mAP reached \(67.6\%~(+1.3\%)\). Little improvements were observed on nuImages-10 \(47.6\%~(+0.1\%)\), but small momentum values led to small but consistent improvements for the more heterogeneous nuImages-23 \(28.5\%~(+0.3\%)\). However, large increases in update volume resulting from choosing a high momentum factor, such as 0.9, significantly disrupted the training process and led to inaccurate predictions, especially on non-IID data. A qualitative comparison between FedAvg and FedOptM is shown in Fig.~\ref{fig_9}.

\subsection{Communication costs and inference speed}

We measured the testing accuracy and inference speed of different YOLOv7 variants, as well as the communication overhead resulting from server-client communications during training, on KITTI, nuImages-10 and nuImages-23. We considered the three P5 models (YOLOv7-tiny, YOLOv7, YOLOv7-X) and the base P6 model (YOLOv7-W6), and we report our findings in Table~\ref{tab:table2}. The federated algorithm was FedOptM, and we reused the server hyperparameters that produced the best performances during ablation. For YOLOv7-W6, letterbox resizing was increased from 640px to 1280px, while the training batch size was reduced from 32 to 8. Training lasted 30 communication rounds of 5 epochs each, and all other hyperparameters remained as previously established for all model variants. To facilitate comparisons, latency measurements were conducted in a single Colab environment separately from model training, and results were averaged over 20 and 5 runs for KITTI and nuImages, respectively. Similarly to the original benchmarking of YOLOv7, we measured latencies on a V100 in FPS at batch size 1, following re-parameterization and model tracing, but without porting the models to ONNX or TensorRT. Only model inference is considered, thus the pre-processing and post-processing costs (including NMS) are not accounted for. The communication cost is reported in megabytes for exactly one-to-one transmission of the symmetrically encrypted learnable parameters between the server and a client, and during a round, there are twice as many transmissions as there are participating clients. The initial broadcast featuring the entire checkpoint of the model has a marginally higher cost than the one reported in the table. Using half-precision during model transfer reduced the communication overhead and is straightforward to implement, and we leave the implementation of more advanced model compression techniques to future work.

\begin{table}[!t]
\caption{General performances for several YOLOv7 variants \label{tab:table2}}
\centering
\begin{tabular}{cccccc}
\toprule
Dataset & Model & Size & mAP & FPS & Overhead \\
\midrule
\multirow{4}{*}{KITTI} 
& YOLOv7-tiny & 640 & 53.4\% & 208 & 12.2MB \\
& YOLOv7 & 640 & 67.6\% & 142 & 74.8MB \\
& YOLOv7-X & 640 & 68.6\% & 123 & 142.1MB \\
& YOLOv7-W6 & 1280 & 71.8\% & 115 & 162.5MB \\
\midrule
\multirow{4}{*}{nuImages-10}
& YOLOv7-tiny & 640 & 33.3\% & 201 & 12.2MB \\
& YOLOv7 & 640 & 47.6\% & 138 & 74.8MB \\
& YOLOv7-X & 640 & 49.4\% & 113 & 142.1MB \\
& YOLOv7-W6 & 1280 & 53.1\% & 92 & 162.6MB \\
\midrule
\multirow{4}{*}{nuImages-23} 
& YOLOv7-tiny & 640 & 18.1\% & 198 & 12.3MB \\
& YOLOv7 & 640 & 28.5\% & 136 & 74.9MB \\
& YOLOv7-X & 640 & 29.6\% & 112 & 142.3MB \\
& YOLOv7-W6 & 1280 & 32.4\% & 92 & 163.0MB \\
\bottomrule
\end{tabular}
\end{table}

\section{Conclusion} \label{conclusion}
In this study, we advocate for FL to address the privacy and scalability challenges inherent to IoV. We propose FedPylot, a practical MPI-based client-server prototype that features hybrid encryption to simulate the federated training of modern object detectors on HPC systems. Through a comprehensive experimental analysis conducted on datasets relevant to autonomous driving, where we considered prediction quality, model latency, and communication overhead, we show the potential of FL to realize object detection and address the real-time processing requirements of autonomous vehicles. Our findings highlight the robustness of the federated optimization of YOLOv7 under realistic heterogeneity constraints, and we hope that FedPylot will invite further research to bring state-of-the-art object detection to FL. Possible future improvements that we are considering for FedPylot include adding support for other object detectors, multimodal sensing capabilities, advanced privacy techniques, asynchronous and low-participation-rate configurations, and model personalization.
\section*{Acknowledgments}
The authors would like to thank the Natural Sciences and Engineering Research Council of Canada, for the financial support of this research. Furthermore, the authors express their gratitude to the BC DRI Group and the Digital Research Alliance of Canada, for the computing resources they provided.

\bibliographystyle{IEEEtran}
\bibliography{IEEEabrv, references}

\begin{thebibliography}{10}
\providecommand{\url}[1]{#1}
\csname url@samestyle\endcsname
\providecommand{\newblock}{\relax}
\providecommand{\bibinfo}[2]{#2}
\providecommand{\BIBentrySTDinterwordspacing}{\spaceskip=0pt\relax}
\providecommand{\BIBentryALTinterwordstretchfactor}{4}
\providecommand{\BIBentryALTinterwordspacing}{\spaceskip=\fontdimen2\font plus
\BIBentryALTinterwordstretchfactor\fontdimen3\font minus \fontdimen4\font\relax}
\providecommand{\BIBforeignlanguage}[2]{{%
\expandafter\ifx\csname l@#1\endcsname\relax
\typeout{** WARNING: IEEEtran.bst: No hyphenation pattern has been}%
\typeout{** loaded for the language `#1'. Using the pattern for}%
\typeout{** the default language instead.}%
\else
\language=\csname l@#1\endcsname
\fi
#2}}
\providecommand{\BIBdecl}{\relax}
\BIBdecl

\bibitem{xuInternetVehiclesBig2018}
W.~Xu \emph{et~al.}, ``Internet of vehicles in big data era,'' \emph{IEEE/CAA J. Autom. Sinica}, vol.~5, no.~1, pp. 19--35, Jan. 2018.

\bibitem{storckSurvey5GTechnology2020}
C.~R. Storck and F.~Duarte-Figueiredo, ``A {survey} of {5G} {technology} {evolution}, {standards}, and {infrastructure} {associated} {with} {vehicle}-to-{everything} {communications} by {Internet} of {Vehicles},'' \emph{IEEE Access}, vol.~8, pp. 117\,593--117\,614, 2020.

\bibitem{grigorescuSurveyDeepLearning2020a}
S.~Grigorescu \emph{et~al.}, ``\BIBforeignlanguage{en}{A survey of deep learning techniques for autonomous driving},'' \emph{\BIBforeignlanguage{en}{J. Field Robot.}}, vol.~37, no.~3, pp. 362--386, 2020.

\bibitem{luBlockchainEmpoweredAsynchronous2020}
Y.~Lu \emph{et~al.}, ``Blockchain empowered asynchronous federated learning for secure data sharing in {Internet} of {Vehicles},'' \emph{{IEEE} Trans. Veh. Technol.}, vol.~69, no.~4, pp. 4298--4311, Apr. 2020.

\bibitem{manias_making_2021}
D.~M. Manias and A.~Shami, ``Making a {case} for {federated} {learning} in the {Internet} of {Vehicles} and {intelligent} {transportation} {systems},'' \emph{{IEEE} Netw.}, vol.~35, no.~3, pp. 88--94, May 2021.

\bibitem{zhou_two-layer_2021}
X.~Zhou \emph{et~al.}, ``Two-{layer} {federated} {learning} {with} {heterogeneous} {model} {aggregation} for {6G} {supported} {Internet} of {Vehicles},'' \emph{{IEEE} Trans. Veh. Technol.}, vol.~70, no.~6, pp. 5308--5317, Jun. 2021.

\bibitem{janai_computer_2020}
J.~Janai \emph{et~al.}, ``\BIBforeignlanguage{English}{Computer {vision} for {autonomous} {vehicles}: {Problems}, {datasets} and {state} of the {art}},'' \emph{\BIBforeignlanguage{English}{Found. Trends Comput. Graph. Vis.}}, vol.~12, no. 1–3, pp. 1--308, Jul. 2020.

\bibitem{xieEfficientFederatedLearning2022}
K.~Xie \emph{et~al.}, ``Efficient federated learning with spike neural networks for traffic sign recognition,'' \emph{{IEEE} Trans. Veh. Technol.}, vol.~71, no.~9, pp. 9980--9992, Sep. 2022.

\bibitem{padariaTrafficSignClassification2023}
A.~A. Padaria \emph{et~al.}, ``Traffic sign classification for autonomous vehicles using split and federated learning underlying {5G},'' \emph{IEEE Open J. Veh. Technol.}, vol.~4, pp. 877--892, 2023.

\bibitem{lianTrafficSignRecognition2024}
Z.~Lian \emph{et~al.}, ``Traffic sign recognition using optimized federated learning in {Internet} of {Vehicles},'' \emph{{IEEE} Internet Things J.}, vol.~11, no.~4, pp. 6722--6729, Feb. 2024.

\bibitem{yuanFedRDPrivacypreservingAdaptive2021}
Y.~Yuan \emph{et~al.}, ``{FedRD}: Privacy-preserving adaptive federated learning framework for intelligent hazardous road damage detection and warning,'' \emph{Future Gener. Comput. Syst.}, vol. 125, Dec. 2021.

\bibitem{alshammari3PodFederatedLearningbased2022}
S.~Alshammari and S.~Song, ``{3Pod}: Federated learning-based 3 dimensional pothole detection for smart transportation,'' in \emph{Proc. {IEEE} {Int.} {Smart} {Cities} {Conf.} ({ISC2})}, Sep. 2022, pp. 1--7.

\bibitem{sahaFederatedLearningBased}
P.~K. Saha \emph{et~al.}, ``\BIBforeignlanguage{en}{Federated learning–based global road damage detection},'' \emph{\BIBforeignlanguage{en}{Comput.-Aided Civil Infrastruct. Eng.}}, 2024.

\bibitem{fantauzzo_feddrive_2022}
L.~Fantauzzo \emph{et~al.}, ``{FedDrive}: Generalizing federated learning to semantic segmentation in autonomous driving,'' in \emph{Proc. IEEE/RSJ Int. Conf. Intell. Robots Syst.}, Oct. 2022, pp. 11\,504--11\,511.

\bibitem{fani_feddrive_2023}
\BIBentryALTinterwordspacing
E.~Fanì \emph{et~al.}, ``{FedDrive} v2: {An} {analysis} of the {impact} of {label} {skewness} in {federated} {semantic} {segmentation} for {autonomous} {driving},'' Oct. 2023. [Online]. Available: \url{http://arxiv.org/abs/2309.13336}
\BIBentrySTDinterwordspacing

\bibitem{rjoub_improving_2021}
G.~Rjoub \emph{et~al.}, ``Improving autonomous vehicles safety in snow weather using federated {YOLO} {CNN} learning,'' in \emph{Int. Conf. Mobile Web Intell. Inform. Syst.}, 2021, pp. 121--134.

\bibitem{chen_federated_2021}
Y.~Chen \emph{et~al.}, ``Federated learning with infrastructure resource limitations in vehicular object detection,'' in \emph{Proc. IEEE/ACM Symp. Edge Comput.}, Dec. 2021, pp. 366--370.

\bibitem{bommel_active_2021}
\BIBentryALTinterwordspacing
J.~R.~v. Bommel, ``\BIBforeignlanguage{en}{Active {learning} during {federated} {learning} for {object} {detection}},'' Jul. 2021. [Online]. Available: \url{https://essay.utwente.nl/86855/}
\BIBentrySTDinterwordspacing

\bibitem{rjoub_active_2022}
G.~Rjoub \emph{et~al.}, ``\BIBforeignlanguage{en}{Active federated {YOLOR} model for enhancing autonomous vehicles safety},'' in \emph{\BIBforeignlanguage{en}{Mobile Web Intell. Inform. Syst.}}, 2022, pp. 49--64.

\bibitem{dai_online_2023}
S.~Dai \emph{et~al.}, ``Online federated learning based object detection across autonomous vehicles in a virtual world,'' in \emph{Proc. IEEE 20th Consum. Commun. Netw. Conf.}, Jan. 2023, pp. 919--920.

\bibitem{wangFederatedDeepLearning2023}
S.~Wang \emph{et~al.}, ``Federated deep learning meets autonomous vehicle perception: Design and verification,'' \emph{IEEE Netw.}, vol.~37, no.~3, pp. 16--25, May 2023.

\bibitem{chi_federated_2023}
F.~Chi \emph{et~al.}, ``Federated semi-supervised learning for object detection in autonomous driving,'' in \emph{Proc. IEEE Int. Conf. Acoust. Speech Signal Process.}, Jun. 2023, pp. 1--5.

\bibitem{rao_sparse_2023}
L.~Rao \emph{et~al.}, ``Sparse federated training of object detection in the {Internet} of {Vehicles},'' in \emph{IEEE Int. Conf. Commun.}, May 2023, pp. 1768--1773.

\bibitem{su_cross-domain_2023}
S.~Su \emph{et~al.}, ``Cross-domain federated object detection,'' in \emph{IEEE Int. Conf. Multimedia Expo}, Jul. 2023, pp. 1469--1474.

\bibitem{kim_navigating_2023}
T.~Kim \emph{et~al.}, ``\BIBforeignlanguage{en}{Navigating data heterogeneity in federated learning: A semi-supervised approach for object detection},'' in \emph{\BIBforeignlanguage{en}{37th Conf. Neural Inf. Process. Syst.}}, Nov. 2023.

\bibitem{zheng_autofed_2023}
T.~Zheng \emph{et~al.}, ``{AutoFed}: Heterogeneity-aware federated multimodal learning for robust autonomous driving,'' in \emph{Proc. 29th Annu. Int. Conf. Mobile Comput. Netw.}\hskip 1em plus 0.5em minus 0.4em\relax ACM, Jul. 2023, no.~15, pp. 1--15.

\bibitem{mishra_swarm_2023}
A.~Mishra \emph{et~al.}, ``Swarm {learning} {in} {autonomous} {driving}: {A} {privacy} {preserving} {approach},'' in \emph{Proc. 15th Int. Conf. Comput. Model. Simul.}, Aug. 2023, pp. 271--277.

\bibitem{chi_federated_2023-1}
F.~Chi \emph{et~al.}, ``Federated cooperative {3D} object detection for autonomous driving,'' in \emph{Proc. IEEE 33rd Int. Workshop Mach. Learn. Signal Process.}, Sep. 2023, pp. 1--6.

\bibitem{khalilFederatedLearningHeterogeneous2024}
\BIBentryALTinterwordspacing
A.~Khalil \emph{et~al.}, ``Federated learning with heterogeneous data handling for robust vehicular object detection,'' May 2024. [Online]. Available: \url{http://arxiv.org/abs/2405.01108}
\BIBentrySTDinterwordspacing

\bibitem{khalilDrivingEfficiencyAdaptive2024}
------, ``Driving towards efficiency: Adaptive resource-aware clustered federated learning in vehicular networks,'' \emph{22nd Mediterranean Commun. Comput. Netw. Conf. IEEE}, Apr. 2024.

\bibitem{jallepalli_federated_2021}
D.~Jallepalli \emph{et~al.}, ``Federated learning for object detection in autonomous vehicles,'' in \emph{Proc. IEEE 7th Int. Conf. Big Data Comput. Service Appl.}, Aug. 2021, pp. 107--114.

\bibitem{hanFederatedLearningbasedTrajectory2022}
M.~Han \emph{et~al.}, ``\BIBforeignlanguage{en}{Federated learning-based trajectory prediction model with privacy preserving for intelligent vehicle},'' \emph{\BIBforeignlanguage{en}{Int. J. Intell. Syst.}}, vol.~37, no.~12, pp. 10\,861--10\,879, 2022.

\bibitem{yuPersonalizedDrivingAssistance2022}
R.~Yu \emph{et~al.}, ``Personalized driving assistance algorithms: Case study of federated learning based forward collision warning,'' \emph{Accident Anal. Prevent.}, vol. 168, p. 106609, Apr. 2022.

\bibitem{wang_yolov7_2023}
C.-Y. Wang \emph{et~al.}, ``\BIBforeignlanguage{en}{{YOLOv7}: {Trainable} {bag}-of-{freebies} {sets} {new} {state}-of-the-{art} for {real}-{time} {object} {detectors}},'' in \emph{\BIBforeignlanguage{en}{Proc. IEEE/CVF Conf. Comput. Vis. Pattern Recognit.}}, Jun. 2023, pp. 7464--7475.

\bibitem{tervenComprehensiveReviewYOLO2023a}
J.~Terven \emph{et~al.}, ``\BIBforeignlanguage{en}{A comprehensive review of {YOLO} architectures in computer vision: From {YOLOv1} to {YOLOv8} and {YOLO}-{NAS}},'' \emph{\BIBforeignlanguage{en}{Mach. Learn. Knowl. Extr.}}, vol.~5, no.~4, pp. 1680--1716, Dec. 2023.

\bibitem{he_fedml_2020}
\BIBentryALTinterwordspacing
C.~He \emph{et~al.}, ``{FedML}: {A} {research} {library} and {benchmark} for {federated} {machine} {learning},'' Nov. 2020. [Online]. Available: \url{http://arxiv.org/abs/2007.13518}
\BIBentrySTDinterwordspacing

\bibitem{zillerPySyftLibraryEasy2021}
A.~Ziller \emph{et~al.}, ``\BIBforeignlanguage{en}{{PySyft}: A library for easy federated learning},'' in \emph{\BIBforeignlanguage{en}{Federated {Learn.} {Syst.}}}\hskip 1em plus 0.5em minus 0.4em\relax Springer, 2021, pp. 111--139.

\bibitem{beutelFlowerFriendlyFederated2022a}
\BIBentryALTinterwordspacing
D.~J. Beutel \emph{et~al.}, ``Flower: A friendly federated learning research framework,'' Mar. 2022. [Online]. Available: \url{http://arxiv.org/abs/2007.14390}
\BIBentrySTDinterwordspacing

\bibitem{mcmahan_ce_2017}
B.~McMahan \emph{et~al.}, ``\BIBforeignlanguage{en}{Communication-efficient learning of deep networks from decentralized data},'' in \emph{\BIBforeignlanguage{en}{Proc. 20th Int. Conf. Artif. Intell. Stat.}}, Apr. 2017, pp. 1273--1282.

\bibitem{reddi_adaptive_2021}
S.~J. Reddi \emph{et~al.}, ``\BIBforeignlanguage{en}{Adaptive federated optimization},'' in \emph{\BIBforeignlanguage{en}{Proc. Int. Conf. Learn. Representations}}, 2021.

\bibitem{tan_towards_2023}
A.~Z. Tan \emph{et~al.}, ``Towards personalized federated learning,'' \emph{{IEEE} Trans. Neural Netw. Learn. Syst.}, vol.~34, no.~12, pp. 9587--9603, Dec. 2023.

\bibitem{nguyen_federated_2021}
D.~C. Nguyen \emph{et~al.}, ``Federated learning meets blockchain in edge computing: Opportunities and challenges,'' \emph{{IEEE} Internet Things J.}, vol.~8, no.~16, pp. 12\,806--12\,825, Aug. 2021.

\bibitem{yang_federated_2019}
Q.~Yang \emph{et~al.}, ``Federated {machine} {learning}: {Concept} and {applications},'' \emph{ACM Trans. Intell. Syst. Technol.}, vol.~10, no.~2, pp. 12:1--12:19, Jan. 2019.

\bibitem{kairouz_advances_2021}
P.~Kairouz \emph{et~al.}, ``\BIBforeignlanguage{English}{Advances and {open} {problems} in {federated} {learning}},'' \emph{\BIBforeignlanguage{English}{Found. Trends Mach. Learn.}}, vol.~14, no. 1–2, pp. 1--210, Jun. 2021.

\bibitem{hsu_measuring_2019}
\BIBentryALTinterwordspacing
T.-M.~H. Hsu \emph{et~al.}, ``Measuring the effects of non-identical data distribution for federated visual classification,'' Sep. 2019. [Online]. Available: \url{http://arxiv.org/abs/1909.06335}
\BIBentrySTDinterwordspacing

\bibitem{li_federated_2020}
T.~Li \emph{et~al.}, ``Federated {optimization} in {heterogeneous} {networks},'' \emph{Proc. Mach. Learn. Syst.}, pp. 429--450, Mar. 2020.

\bibitem{karimireddy_scaffold_2020}
S.~P. Karimireddy \emph{et~al.}, ``{SCAFFOLD}: {Stochastic} {controlled} {averaging} for {federated} {learning},'' in \emph{Proc. 37th Int. Conf. Mach. Learn.}, Nov. 2020, pp. 5132--5143.

\bibitem{li_model-contrastive_2021}
Q.~Li \emph{et~al.}, ``\BIBforeignlanguage{en}{Model-{contrastive} {federated} {learning}},'' in \emph{\BIBforeignlanguage{en}{Proc. IEEE/CVF Conf. Comput. Vis. Pattern Recognit.}}, Jun. 2021, pp. 10\,713--10\,722.

\bibitem{miaoFedSegClassHeterogeneousFederated2023b}
J.~Miao \emph{et~al.}, ``{FedSeg}: Class-heterogeneous federated learning for semantic segmentation,'' in \emph{Proc. IEEE/CVF Conf. Comput. Vis. Pattern Recognit.}, Jun. 2023, pp. 8042--8052.

\bibitem{kingma_adam_2017}
\BIBentryALTinterwordspacing
D.~P. Kingma and J.~Ba, ``Adam: {A} {method} for {stochastic} {optimization},'' Jan. 2017. [Online]. Available: \url{http://arxiv.org/abs/1412.6980}
\BIBentrySTDinterwordspacing

\bibitem{liu_accelerating_2020}
W.~Liu \emph{et~al.}, ``Accelerating {federated} {learning} via {momentum} {gradient} {descent},'' \emph{{IEEE} Trans. Parallel Distrib. Syst.}, vol.~31, no.~8, pp. 1754--1766, Aug. 2020.

\bibitem{xu_fedcm_2021}
\BIBentryALTinterwordspacing
J.~Xu \emph{et~al.}, ``{FedCM}: {Federated} {learning} with {client}-level {momentum},'' Jun. 2021. [Online]. Available: \url{http://arxiv.org/abs/2106.10874}
\BIBentrySTDinterwordspacing

\bibitem{chenImportanceApplicabilityPreTraining2022}
H.-Y. Chen \emph{et~al.}, ``\BIBforeignlanguage{en}{On the importance and applicability of pre-training for federated learning},'' in \emph{\BIBforeignlanguage{en}{Proc. Int. Conf. Learn. Representations}}, Sep. 2022.

\bibitem{nguyenWhereBeginImpact2022}
J.~Nguyen \emph{et~al.}, ``\BIBforeignlanguage{en}{Where to begin? {On} the impact of pre-training and initialization in federated learning},'' in \emph{\BIBforeignlanguage{en}{Proc. Int. Conf. Learn. Representations}}, Sep. 2022.

\bibitem{yao_how_1986}
A.~C.-C. Yao, ``How to generate and exchange secrets,'' in \emph{Proc. 27th Annu. Symp. Found. Comput. Sci. (FOCS)}, Oct. 1986, pp. 162--167.

\bibitem{gentry_fully_2009}
C.~Gentry, ``Fully homomorphic encryption using ideal lattices,'' in \emph{Proc. 41st Annu. ACM Symp. Theory Comput. (STOC)}, May 2009, pp. 169--178.

\bibitem{dwork_calibrating_2006}
C.~Dwork \emph{et~al.}, ``\BIBforeignlanguage{en}{Calibrating {noise} to {sensitivity} in {private} {data} {analysis}},'' in \emph{\BIBforeignlanguage{en}{Proc. Theory Cryptogr. Conf.}}, 2006, pp. 265--284.

\bibitem{zaidi_survey_2022}
S.~S.~A. Zaidi \emph{et~al.}, ``A survey of modern deep learning based object detection models,'' \emph{Digit. Signal Process.}, vol. 126, p. 103514, Jun. 2022.

\bibitem{girshick_rich_2014}
R.~Girshick \emph{et~al.}, ``Rich feature hierarchies for accurate object detection and semantic segmentation,'' in \emph{Proc. IEEE Conf. Comput. Vis. Pattern Recognit.}, Jun. 2014, pp. 580--587.

\bibitem{he_spatial_2015}
K.~He \emph{et~al.}, ``Spatial pyramid pooling in deep convolutional networks for visual recognition,'' \emph{{IEEE} Trans. Pattern Anal. Mach. Intell.}, vol.~37, no.~9, pp. 1904--1916, Sep. 2015.

\bibitem{redmon_you_2016}
J.~Redmon \emph{et~al.}, ``You only look once: Unified, real-time object detection,'' in \emph{Proc. IEEE Conf. Comput. Vis. Pattern Recognit.}, Jun. 2016, pp. 779--788.

\bibitem{liu_ssd_2016}
W.~Liu \emph{et~al.}, ``{SSD}: Single shot {MultiBox} detector,'' in \emph{Proc. Eur. Conf. Comput. Vis.}, 2016, pp. 21--37.

\bibitem{linFocalLossDense2017}
T.-Y. Lin \emph{et~al.}, ``Focal loss for dense object detection,'' in \emph{Proc. IEEE Int. Conf. on Comput. Vis.}, 2017, pp. 2980--2988.

\bibitem{carion_end--end_2020}
N.~Carion \emph{et~al.}, ``End-to-end object detection with transformers,'' in \emph{Proc. Eur. Conf. Comput. Vis.}, 2020, pp. 213--229.

\bibitem{lv_detrs_2023}
\BIBentryALTinterwordspacing
W.~Lv \emph{et~al.}, ``{DETRs} beat {YOLOs} on real-time object detection,'' Jul. 2023. [Online]. Available: \url{http://arxiv.org/abs/2304.08069}
\BIBentrySTDinterwordspacing

\bibitem{padilla_survey_2020}
R.~Padilla \emph{et~al.}, ``A survey on performance metrics for object-detection algorithms,'' in \emph{Proc. Int. Conf. Syst. Signals Image Process}, Jul. 2020, pp. 237--242.

\bibitem{Jocher_YOLOv5_by_Ultralytics_2020}
\BIBentryALTinterwordspacing
G.~Jocher, ``{YOLOv5 by Ultralytics},'' May 2020. [Online]. Available: \url{https://github.com/ultralytics/yolov5}
\BIBentrySTDinterwordspacing

\bibitem{lin_microsoft_2014}
T.-Y. Lin \emph{et~al.}, ``\BIBforeignlanguage{en}{Microsoft {COCO}: {Common} {objects} in {context}},'' in \emph{\BIBforeignlanguage{en}{Proc. Eur. Conf. Comput. Vis.}}, 2014, pp. 740--755.

\bibitem{wangYouOnlyLearn2023}
C.-Y. Wang \emph{et~al.}, ``You {only} {learn} {one} {representation}: {Unified} {network} for {multiple} {tasks},'' \emph{J. Inf. Sci. Eng.}, vol.~39, no.~3, pp. 691--709, May 2023.

\bibitem{bochkovskiy_yolov4_2020}
\BIBentryALTinterwordspacing
A.~Bochkovskiy \emph{et~al.}, ``{YOLOv4}: Optimal speed and accuracy of object detection,'' Apr. 2020. [Online]. Available: \url{http://arxiv.org/abs/2004.10934}
\BIBentrySTDinterwordspacing

\bibitem{wangDesigningNetworkDesign2022a}
\BIBentryALTinterwordspacing
C.-Y. Wang \emph{et~al.}, ``Designing network design strategies through gradient path analysis,'' Nov. 2022. [Online]. Available: \url{http://arxiv.org/abs/2211.04800}
\BIBentrySTDinterwordspacing

\bibitem{lin_feature_2017}
T.-Y. Lin \emph{et~al.}, ``Feature {pyramid} {networks} for {object} {detection},'' in \emph{Proc. IEEE Conf. Comput. Vis. Pattern Recognit.}, 2017, pp. 2117--2125.

\bibitem{zheng_distance-iou_2020}
Z.~Zheng \emph{et~al.}, ``Distance-{IoU} {loss}: {Faster} and {better} {learning} for {bounding} {box} {regression},'' \emph{Proc. AAAI Conf. Artif. Intell.}, pp. 12\,993--13\,000, Apr. 2020.

\bibitem{geYOLOXExceedingYOLO2021a}
\BIBentryALTinterwordspacing
Z.~Ge \emph{et~al.}, ``{YOLOX}: {Exceeding} {YOLO} {series} in 2021,'' Aug. 2021. [Online]. Available: \url{http://arxiv.org/abs/2107.08430}
\BIBentrySTDinterwordspacing

\bibitem{li_yolov6_2023}
\BIBentryALTinterwordspacing
C.~Li \emph{et~al.}, ``{YOLOv6} v3.0: A full-scale reloading,'' Jan. 2023. [Online]. Available: \url{http://arxiv.org/abs/2301.05586}
\BIBentrySTDinterwordspacing

\bibitem{Jocher_Ultralytics_YOLO_2023}
\BIBentryALTinterwordspacing
G.~Jocher \emph{et~al.}, ``{Ultralytics YOLO},'' Jan. 2023. [Online]. Available: \url{https://github.com/ultralytics/ultralytics}
\BIBentrySTDinterwordspacing

\bibitem{wangYOLOv9LearningWhat2024}
\BIBentryALTinterwordspacing
C.-Y. Wang \emph{et~al.}, ``{YOLOv9}: Learning what you want to learn using programmable gradient information,'' Feb. 2024. [Online]. Available: \url{http://arxiv.org/abs/2402.13616}
\BIBentrySTDinterwordspacing

\bibitem{redmonYOLO9000BetterFaster2017a}
J.~Redmon and A.~Farhadi, ``{YOLO9000}: Better, faster, stronger,'' in \emph{Proc. IEEE Conf. Comput. Vis. Pattern Recognit.}, 2017, pp. 7263--7271.

\bibitem{renFasterRCNNRealTime2017b}
S.~Ren \emph{et~al.}, ``Faster {R}-{CNN}: Towards real-time object detection with region proposal networks,'' \emph{{IEEE} Trans. Pattern Anal. Mach. Intell.}, vol.~39, no.~6, pp. 1137--1149, Jun. 2017.

\bibitem{redmonYOLOv3IncrementalImprovement2018a}
\BIBentryALTinterwordspacing
J.~Redmon and A.~Farhadi, ``{YOLOv3}: An incremental improvement,'' Apr. 2018. [Online]. Available: \url{http://arxiv.org/abs/1804.02767}
\BIBentrySTDinterwordspacing

\bibitem{simonComplexYOLOEulerRegionProposalRealTime2019}
M.~Simon \emph{et~al.}, ``Complex-{YOLO}: An {Euler}-region-proposal for real-time {3D} object detection on point clouds,'' in \emph{Proc. Eur. Conf. Comput. Vis. Workshops}, 2019, pp. 197--209.

\bibitem{paszke_pytorch_2019}
A.~Paszke \emph{et~al.}, ``{PyTorch}: {An} {imperative} {style}, {high}-{performance} {deep} {learning} {library},'' in \emph{Proc. Adv. Neural Inf. Process. Syst.}, 2019, pp. 8026--8037.

\bibitem{birrittellaIntelOmnipathArchitecture2015}
M.~S. Birrittella \emph{et~al.}, ``Intel {Omni}-{Path} {Architecture}: Enabling scalable, high performance fabrics,'' in \emph{Proc. IEEE 23rd Annu. Symp. High-Perform. Interconnects}, Aug. 2015, pp. 1--9.

\bibitem{gabriel_open_2004}
E.~Gabriel \emph{et~al.}, ``\BIBforeignlanguage{en}{Open {MPI}: Goals, concept, and design of a next generation {MPI} implementation},'' in \emph{\BIBforeignlanguage{en}{Proc. 11th Eur. PVM/MPI Users’ Group Meeting}}, Budapest, Hungary, Sep. 2004, pp. 97--104.

\bibitem{dalcin_mpi4py_2021}
L.~Dalcin and Y.-L.~L. Fang, ``mpi4py: Status update after 12 years of development,'' \emph{{IEEE} Comput. Sci. Eng.}, vol.~23, no.~4, pp. 47--54, Jul. 2021.

\bibitem{rivest_method_1978}
R.~L. Rivest \emph{et~al.}, ``A method for obtaining digital signatures and public-key cryptosystems,'' \emph{Commun. ACM}, vol.~21, no.~2, pp. 120--126, Feb. 1978.

\bibitem{bellareOptimalAsymmetricEncryption1995}
M.~Bellare and P.~Rogaway, ``\BIBforeignlanguage{en}{Optimal asymmetric encryption},'' in \emph{\BIBforeignlanguage{en}{Proc. {EUROCRYPT}'94}}, 1995, pp. 92--111.

\bibitem{daemen_aes_1999}
\BIBentryALTinterwordspacing
J.~Daemen and V.~Rijmen, ``\BIBforeignlanguage{en}{{AES} proposal: {Rijndael}},'' 1999. [Online]. Available: \url{https://www.cs.miami.edu/home/burt/learning/Csc688.012/rijndael/rijndael_doc_V2.pdf}
\BIBentrySTDinterwordspacing

\bibitem{mcgrew_galoiscounter_2004}
\BIBentryALTinterwordspacing
D.~McGrew and J.~Viega, ``\BIBforeignlanguage{en}{The {Galois}/counter mode of operation ({GCM})},'' \emph{\BIBforeignlanguage{en}{NIST Modes of Operation Process}}, 2004. [Online]. Available: \url{https://csrc.nist.rip/groups/ST/toolkit/BCM/documents/proposedmodes/gcm/gcm-spec.pdf}
\BIBentrySTDinterwordspacing

\bibitem{percival_stronger_2009}
\BIBentryALTinterwordspacing
C.~Percival, ``Stronger key derivation via sequential memory-hard functions,'' Jan. 2009. [Online]. Available: \url{https://www.bsdcan.org/2009/schedule/attachments/87_scrypt.pdf}
\BIBentrySTDinterwordspacing

\bibitem{geiger_are_2012}
A.~Geiger \emph{et~al.}, ``Are we ready for autonomous driving? {The} {KITTI} vision benchmark suite,'' in \emph{Proc. IEEE Conf. Comput. Vis. Pattern Recognit.}, Jun. 2012, pp. 3354--3361.

\bibitem{caesar_nuscenes_2020}
H.~Caesar \emph{et~al.}, ``{nuScenes}: A multimodal dataset for autonomous driving,'' in \emph{Proc. IEEE/CVF Conf. Comput. Vis. Pattern Recognit.}, 2020, pp. 11\,621--11\,631.

\end{thebibliography}

\vfill

\end{document}